\documentclass[twoside,11pt]{article}

\usepackage{jmlr2e}
\usepackage[utf8]{inputenc}  
\usepackage{booktabs}        
\usepackage{pifont}          
\usepackage{xspace}
\usepackage{listings}
\usepackage[frozencache,cachedir=.]{minted}
\usepackage{hyperref}
\usepackage{graphicx}

\newcommand{\River}{\textsf{River}\xspace}
\newcommand{\river}{\textsf{river}\xspace}
\newcommand{\creme}{\textsf{Creme}\xspace}
\newcommand{\skmultiflow}{\textsf{scikit-multiflow}\xspace}
\newcommand{\sklearn}{\textsf{scikit-learn}\xspace}

\ShortHeadings{River: online machine learning in Python}{River development team}
\firstpageno{1}

\begin{document}

\title{River: machine learning for streaming data in Python}

\author{\name Jacob Montiel\thanks{Co-first authors.} \email jacob.montiel@waikato.ac.nz \\
       \addr AI Institute, University of Waikato, Hamilton, New Zealand
       \AND
       \name Max Halford\footnotemark[1] \email max.halford@alan.eu \\
       \addr Alan, Paris, France
       \AND
       \name Saulo Martiello Mastelini \email mastelini@usp.br \\
       \addr Institute of Mathematics and Computer Sciences, University of S\~{a}o Paulo, S\~{a}o Carlos, Brazil\
       \AND
       \name Geoffrey Bolmier \email geoffrey.bolmier@volvocars.com \\
       \addr 
Volvo Car Corporation, Göteborg, Sweden
       \AND
       \name Raphael Sourty \email raphael.sourty@irit.fr \\
       \addr IRIT, Université Paul Sabatier, Toulouse, France \\ 
       \addr Renault, Paris, France
       \AND
       \name Robin Vaysse \email robin.vaysse@irit.fr \\
       \addr IRIT, Université Paul Sabatier, Toulouse, France\\
       \addr Octogone Lordat, Université Jean-Jaures, Toulouse, France
       \AND
       \name Adil Zouitine \email adil.zouitine@irt-saintexupery.com \\
       \addr IRT Saint Exupéry, Toulouse, France
       \AND
       \name Heitor Murilo Gomes \email heitor.gomes@waikato.ac.nz \\
       \addr AI Institute, University of Waikato, Hamilton, New Zealand
       \AND
       \name Jesse Read \email jesse.read@polytechnique.edu \\
       \addr LIX, École Polytechnique, Institut Polytechnique de Paris, Palaiseau, France
       \AND
       \name Talel Abdessalem \email talel.abdessalem@telecom-paris.fr \\
       \addr LTCI, Télécom Paris, Institut Polytechnique de Paris, Palaiseau, France
       \AND
       \name Albert Bifet \email abifet@waikato.ac.nz \\
       \addr AI Institute, University of Waikato, Hamilton, New Zealand\\
       \addr LTCI, Télécom Paris, Institut Polytechnique de Paris, Palaiseau, France
       }

\editor{TBD}

\maketitle

\begin{abstract}

\River is a machine learning library for dynamic data streams and continual learning. It provides multiple state-of-the-art learning methods, data generators/transformers, performance metrics and evaluators for different stream learning problems. It is the result from the merger of the two most popular packages for stream learning in Python: \creme and \skmultiflow. \River introduces a revamped architecture based on the lessons learnt from the seminal packages. \River's ambition is to be the go-to library for doing machine learning on streaming data. Additionally, this open source package brings under the same umbrella a large community of practitioners and researchers. The source code is available at \href{https://github.com/online-ml/river}{https://github.com/online-ml/river}.
\end{abstract}


\section{Introduction}
In machine learning, the conventional approach is to process data in batches or chunks. Batch learning models assume that all the data is available at once. When a new batch of data is available, said models have to be retrained from scratch. The assumption of data availability is a hard constraint for the application of machine learning in multiple real-world applications where data is continuously generated. Additionally, storing historical data requires dedicated storage and processing resources which in some cases might be impractical, e.g. storing the network logs from a data center. A different approach is to treat data as a stream, in other words, as an infinite sequence of items; data is not stored and models continuously learn one data sample at a time \citep{MOA-book}.

\creme \citep{creme} and \skmultiflow \citep{skmultiflow} are two open-source libraries to perform machine learning in the stream setting. These original libraries started as independent projects with the same goal, to provide to the community the tools to advance the state of streaming machine learning and promote its usage on real-world applications. \River is the merger of these projects, combining the strengths of both projects while leveraging the lessons learnt from their development. \River is mainly written in Python, with some core elements written in Cython \citep{Cython} for performance.

Supported applications of \river are generally as diverse as those found in traditional batch settings, including: classification, regression, clustering and representation learning, multi-label and multi-output learning, forecasting, and anomaly detection. 

\section{Architecture}
Machine learning models in \river are extended classes of specialized \mintinline{python}{mixins} depending on the learning task, e.g. classification, regression, clustering, etc. This ensures compatibility across the library and eases the extension/modification of existing models and the creation of new models compatible with \river.

All predictive models perform two core functions: learn and predict. Learning takes place in the \mintinline{python}{learn_one} method (updates the internal state of the model). Depending on the learning task, models provide predictions via the \mintinline{python}{predict_one} (classification, regression, and clustering), \mintinline{python}{predict_proba_one} (classification), and \mintinline{python}{score_one} (anomaly detection) methods. Note that \river also contains transformers, which are stateful objects that transform an input via a \mintinline{python}{transform_one} method.

In the following example, we show a complete machine learning task (learning, prediction and performance measurement) easily implemented in a couple lines of code:

\begin{minted}[
frame=lines,
framesep=2mm,
baselinestretch=1.2,
fontsize=\footnotesize
]{python}
>>> from river import evaluate
>>> from river import metrics
>>> from river import synth
>>> from river import tree

>>> stream = synth.Waveform(seed=42).take(1000)
>>> model = tree.HoeffdingTreeClassifier()
>>> metric = metrics.Accuracy()

>>> evaluate.progressive_val_score(stream, model, metric)
Accuracy: 77.58%
\end{minted}

\subsection{Why dictionaries?}
The de facto container for \textit{multidimensional}, homogeneous arrays of fixed-size items in Python is the \mintinline{python}{numpy.ndarray} \citep{Numpy}. However, in the stream setting, data is available one sample at a time. Dictionaries are an efficient way to store \textit{one-dimensional} data with $O(1)$ lookup and insertion \citep{high_performance_python}\footnote{The actual performance of this operations can be affected by the size of the data to store. We assume that samples from a data stream are relatively small.}. Additional advantages of dictionaries include:
\begin{itemize}
\itemsep0em
    \item Accessing data by name rather than by position is convenient from a user perspective.
    \item The ability to store different data types. For instance, the categories of a nominal feature can be encoded as strings alongside numeric features.
    \item The flexibility to handle new features that might appear in the stream (feature evolution) and sparse data.
\end{itemize}

\River provides an efficient Cython-based extension of dictionary structures that supports operations commonly applied to unidimensional arrays. These operations include, for instance, the four basic algebraic operations, exponentiation, and the dot product.

\subsection{Pipelines}
Pipelines are an integral part of \river. They are a convenient and elegant way to ``chain'' a sequence of operations and warrant reproducibility. A pipeline is essentially a list of estimators that are applied in sequence. The only requirement is that the first $n - 1$ steps are transformers. The last step can be a regressor, a classifier, a clusterer, a transformer, etc. For example, some models such as logistic regression are sensitive to the scale of the data. A best practice is to scale the data before feeding it to a linear model. We can chain the scaler transformer with a logistic regression model via a \mintinline{python}{|} (pipe) operator as follows:

\begin{minted}[
frame=lines,
framesep=2mm,
baselinestretch=1.2,
fontsize=\footnotesize
]{python}
>>> from river import linear_model
>>> from river import preprocessing

>>> model = (preprocessing.StandardScaler() |
...          linear_model.LogisticRegression())
\end{minted}

\subsection{Instance-incremental and batch-incremental}
Instance-incremental methods update their internal state one sample at a time. Another approach is to use mini-batches of data, known as batch-incremental learning. \River offers some limited support for batch-incremental methods. Mixins include dedicated methods to process data in mini-batches, designated by the suffix \mintinline{python}{_many} instead of \mintinline{python}{_one}, e.g. \mintinline{python}{learn_one()} -- \mintinline{python}{learn_many()}. These methods expect \mintinline{python}{pandas.DataFrame} \citep{Pandas} as input data, a flexible data structure with labeled axes. This in turn allows a uniform interface for both instance-incremental and batch-incremental learning.

\section{Benchmark}

We benchmark the implementation of 3 ml algorithms\footnote{These are incremental-learning models. \sklearn has many other batch-learning models available. On the other hand, \river includes incremental-learning methods available in \creme and \skmultiflow.}: Gaussian Naive Bayes (\textsf{GNB}), Logistic Regression (\textsf{LR}), and Hoeffding Tree (\textsf{HT}). Table~\ref{tab:benchmark_acc} shows similar accuracy for all models. Table~\ref{tab:benchmark_time} shows the processing time (learn and predict) for the same models where \river models perform at least as fast but overall faster than the rest. Tests are performed on the \textsf{Elec2} dataset~\citep{Elec2} which has 45312 samples with 8 numerical features. Reported processing time is the average of running the experiment 7 times on a system with a 2.4 GHz Quad-Core Intel Core i5 processor and 16GB of RAM. Additional benchmarks for other ml tasks and packages are available in the project's repository\footnote{\href{https://github.com/online-ml/river/tree/master/benchmarks}{https://github.com/online-ml/river/tree/master/benchmarks}}.

\begin{table}[t]
\caption{Benchmark accuracy (\%) for the \textsf{Elec2} dataset.}
\label{tab:benchmark_acc}
\centering
\footnotesize
\begin{tabular}{@{}lcccc@{}}
\toprule
model        & \sklearn & \creme  & \skmultiflow & \River  \\ \midrule
\textsf{GNB} & $73.22$  & $72.87$ & $73.30$      & $72.87$ \\
\textsf{LR}  & $68.01$  & $67.97$ & NA           & $67.97$ \\
\textsf{HT}  & NA       & $74.48$ & $75.82$      & $75.55$ \\ \bottomrule
\end{tabular}
\end{table}
\begin{table}[t]
\caption{Benchmark processing time (seconds) for the \textsf{Elec2} dataset.}
\label{tab:benchmark_time}
\resizebox{\textwidth}{!}{
\centering
\begin{tabular}{@{}lccccccccc@{}}
\toprule
             & \multicolumn{2}{c}{\sklearn}    & \multicolumn{2}{c}{\creme}    & \multicolumn{2}{c}{\skmultiflow} & \multicolumn{2}{c}{\River} \\
             \cmidrule(l){2-3} \cmidrule(l){4-5} \cmidrule(l){6-7} \cmidrule(l){8-9} 
model        & learn          & predict        & learn         & predict       & learn         & predict          & learn         & predict      \\ \cmidrule(l){1-9} 
\textsf{GNB} & $10.94\pm0.26$ & $5.43\pm0.10$  & $0.32\pm0.01$ & $3.22\pm0.09$ & $1.39\pm0.02$ & $2.91\pm0.03$    & $0.32\pm0.01$ & $3.27\pm0.13$ \\
\textsf{LR}  & $8.72\pm0.14$  & $3.15\pm0.06$  & $2.03\pm0.04$ & $0.42\pm0.01$ & NA            & NA               & $0.95\pm0.06$ & $0.18\pm0.01$ \\
\textsf{HT}  & NA             & NA             & $2.66\pm0.06$ & $0.48\pm0.02$ & $2.95\pm0.06$ & $2.21\pm0.03$    & $0.99\pm0.04$ & $0.65\pm0.03$ \\ \bottomrule
\end{tabular}}
\end{table}

\section{Summary}

\River has been developed to satisfy the evolving needs of a major machine learning community -- learning from data streams. The architecture has been designed for both flexibility and ease of use, with the goal of supporting successful use in diverse domains, including in industrial applications as well as in academic research. On benchmark tests performs at least as well as related (but more limited) methods.

\vskip 0.2in
\bibliography{bibliography}

\end{document}